\documentclass[letterpaper, 10 pt, conference]{ieeeconf}  % Comment this line out if you need a4paper

\IEEEoverridecommandlockouts                              % This command is only needed if 
\usepackage{graphics} % for pdf, bitmapped graphics files
\usepackage{epsfig} % for postscript graphics files
\usepackage{times} % assumes new font selection scheme installed
\usepackage{amsmath} % assumes amsmath package installed
\usepackage{amssymb}  % assumes amsmath package installed
\let\proof\relax 
\let\endproof\relax
\usepackage{amsthm}
\usepackage[T1]{fontenc}
\usepackage{soul}
\usepackage{url}
\usepackage[hidelinks]{hyperref}
\usepackage[utf8]{inputenc}
\usepackage[small]{caption}
\usepackage{booktabs}
\usepackage{algorithm}
\usepackage[noend]{algpseudocode}
\usepackage{algorithmicx}
\usepackage[switch]{lineno}
\usepackage{subcaption}
\usepackage{cleveref}

\usepackage[dvipsnames]{xcolor}

\definecolor{mycustomred}{RGB}{204, 10, 10}
\newcommand{\ral}[1]{#1}
\algdef{SE}[SWITCH]{Switch}{EndSwitch}[1]{\algorithmicswitch\ #1\ \algorithmicdo}{\algorithmicend\ \algorithmicswitch}%
\algdef{SE}[CASE]{Case}{EndCase}[1]{\algorithmiccase\ #1}{\algorithmicend\ \algorithmiccase}%
\algtext*{EndSwitch}%
\algtext*{EndCase}%
\algrenewcommand\textproc{}

\algnewcommand\algorithmicinput{\textbf{Input:}}
\algnewcommand\INPUT{\item[\algorithmicinput]}
\algnewcommand\algorithmicoutput{\textbf{Output:}}
\algnewcommand\OUTPUT{\item[\algorithmicoutput]}

\algnewcommand\algorithmicswitch{\textbf{switch}}
\algnewcommand\algorithmiccase{\textbf{case}}
\algnewcommand\algorithmicassert{\texttt{assert}}
\algnewcommand\Assert[1]{\State \algorithmicassert(#1)}%
\algdef{SE}[SWITCH]{Switch}{EndSwitch}[1]{\algorithmicswitch\ #1\ \algorithmicdo}{\algorithmicend\ \algorithmicswitch}%
\algdef{SE}[CASE]{Case}{EndCase}[1]{\algorithmiccase\ #1}{\algorithmicend\ \algorithmiccase}%
\algtext*{EndSwitch}%
\algtext*{EndCase}%

\title{\LARGE \bf
Enhancing Lifelong Multi-Agent Path Finding with Cache Mechanism
}

\author{Yimin Tang$^{1*}$, Zhenghong Yu$^{2*}$, Yi Zheng$^1$, T. K. Satish Kumar$^1$, Jiaoyang Li$^3$, Sven Koenig$^{1,4}$
\thanks{$^{*}$Equal Contribution}%
\thanks{$^{1}$University of Southern California,
        {\tt\small yimintan@usc.edu, yzheng63@usc.edu, tkskwork@gmail.com}}%
\thanks{$^{2}$University of Wisconsin–Madison,
        {\tt\small zyu379@wisc.edu}}%
\thanks{$^{3}$Carnegie Mellon University,
        {\tt\small jiaoyanl@andrew.cmu.edu}}%
\thanks{$^{4}$University of California, Irvine,
        {\tt\small sven.koenig@uci.edu}}%
}%

\begin{document}

\maketitle
\thispagestyle{empty}
\pagestyle{empty}

%%%%%%%%%%%%%%%%%%%%%%%%%%%%%%%%%%%%%%%%%%%%%%%%%%%%%%%%%%%%%%%%%%%%%%%%%%%%%%%%
\begin{abstract}
Multi-Agent Path Finding (MAPF), which focuses on finding collision-free paths for multiple robots, is crucial in autonomous warehouse operations. Lifelong MAPF (L-MAPF), where agents are continuously reassigned new targets upon completing their current tasks, offers a more realistic approximation of real-world warehouse scenarios. While cache storage systems can enhance efficiency and reduce operational costs, existing approaches primarily rely on expectations and mathematical models, often without adequately addressing the challenges of multi-robot planning and execution.
In this paper, we introduce a novel mechanism called Lifelong MAPF with Cache Mechanism (L-MAPF-CM), which integrates high-level cache storage with low-level path planning. We have involved a new type of map grid called cache for temporary item storage. Additionally, we involved a task assigner (TA) with a locking mechanism to bridge the gap between the new cache grid and L-MAPF algorithm. The TA dynamically allocates target locations to agents based on their status in various scenarios. We evaluated L-MAPF-CM using different cache replacement policies and task distributions. L-MAPF-CM has demonstrated performance improvements particularly with high cache hit rates and smooth traffic conditions. 
\end{abstract}

%%%%%%%%%%%%%%%%%%%%%%%%%%%%%%%%%%%%%%%%%%%%%%%%%%%%%%%%%%%%%%%%%%%%%%%%%%%%%%%%
\section{Introduction}

Automated warehouses, a multibillion-dollar industry led by companies like Amazon, Geekplus and inVia, rely on hundreds of robots to transport goods efficiently~\cite{wurman2008coordinating}. A critical aspect of these operations is planning collision-free paths for robots, a task that can be abstracted as the Multi-Agent Path Finding (MAPF) problem~\cite{stern2019multi}. The MAPF problem requires planning collision-free paths for multiple agents from their start locations to pre-assigned target locations in a known environment while minimizing a specific cost function. Various algorithms have been developed to solve this problem optimally or suboptimally, such as \(M^*\)~\cite{wagner2011m}, Conflict Based Search (CBS)~\cite{sharon2015conflict}, Enhanced CBS~\cite{barer2014suboptimal} and LaCAM~\cite{okumura2023lacam}. 

Although MAPF is classified as an NP-hard problem~\cite{yu2013structure}, it remains a simplified approximation of real-world warehouse planning. It represents a `one-shot' version of the real application challenge, where an agent only needs to reach a single target location and then remains stationary until every agent has arrived at their respective targets. To address this limitation, a more realistic variant called Lifelong MAPF (L-MAPF)~\cite{lifelong2017} has been introduced. In L-MAPF, agents are continuously assigned new targets when they reach their current ones, better reflecting the ongoing nature of warehouse operations. Various algorithms have been developed, such as RHCR~\cite{li2021lifelong}, MAPF-LNS~\cite{li2021anytime} and PIBT~\cite{okumura2022priority}.

There are also works focused on enhancing warehouse efficiency through smart storage strategies. Some methods~\cite{park2001optimal,gagliardi2012storage,li2020storage} focus on optimizing the arrangement of item positions within the warehouse. Other approaches~\cite{YU2008377, POHL2009367, OzakiHOHRO16} involve the use of temporal storage areas (cache), aiming to determine the optimal positions and sizes of these areas. However, these methods primarily rely on expectations and mathematical models without addressing the low-level path planning component.

\ral{In this paper, we are particularly interested in the performance of cache map layout design in the context of real agent path planning, moving beyond the limitations of mathematical expectation models. We present a dynamic cache delivery mechanism named Lifelong MAPF with Cache Mechanism (L-MAPF-CM), inspired by cache design---a foundational idea extensively utilized in computer architecture, databases, and various other domains. This cache mechanism could be adaptable to multiple L-MAPF algorithms, warehouse storage strategies, and incoming task distributions. We introduced a task assigner (TA) to allocate target locations to agents and manage their statuses under various conditions.
In our cache mechanism, determining which agent should go to which location depends on both the cache and agent statuses. Thus, the TA plays a crucial role in making these decisions by overseeing all agents and maintaining the cache status within our framework.}

\ral{Overall, L-MAPF-CM has shown performance improvements across most test settings. The main contributions of this paper are listed as follows: (1) We introduced a cache mechanism in L-MAPF, termed L-MAPF-CM, which integrates cache map layout with low-level agent path planning. To the best of our knowledge, this is the first work to combine cache layout with pathfinding at this level. (2) To ensure compatibility of L-MAPF algorithms with the new cache map layout, we designed a task assigner (TA) to maintain the agents' workflow. The TA is primarily based on a state machine and a lock mechanism to ensure the smooth execution of tasks. (3) We evaluated L-MAPF-CM across various input distributions, numbers of agents and caches, and cache replacement policies. Our experiments show that high cache hit rates and smooth traffic flow are critical to the success of L-MAPF-CM.}

\section{Problem Definition}

Lifelong Multi-Agent Path Finding (L-MAPF) problem presents unique challenges due to the dynamic and unending nature of the tasks. 
Let $I=\{1,2,\cdots,N\}$ denote a set of $N$ agents. $G = (V,E)$ represents an undirected graph, where each vertex $v \in V$ represents a possible location of an agent, and each edge $e \in E$ is a unit-length edge between two vertices, allowing an agent to move from one vertex to another. Self-loop edges are allowed, which represent ``wait-in-place'' actions. Each agent $i\in I$ has a unique start location $s_i \in V$. There is also a target queue \(T=[t_1, t_2, ...]\), where \(t_i \in V\). And the target locations in \(T\) are allocated to agents when agents arrive at their previous target locations. The objective is to minimize a specific cost function---in this paper, we focus on maximizing the throughput, defined as the number of tasks completed per timestep.

\begin{figure}[t!]
\centering
\includegraphics[width=0.48\textwidth]{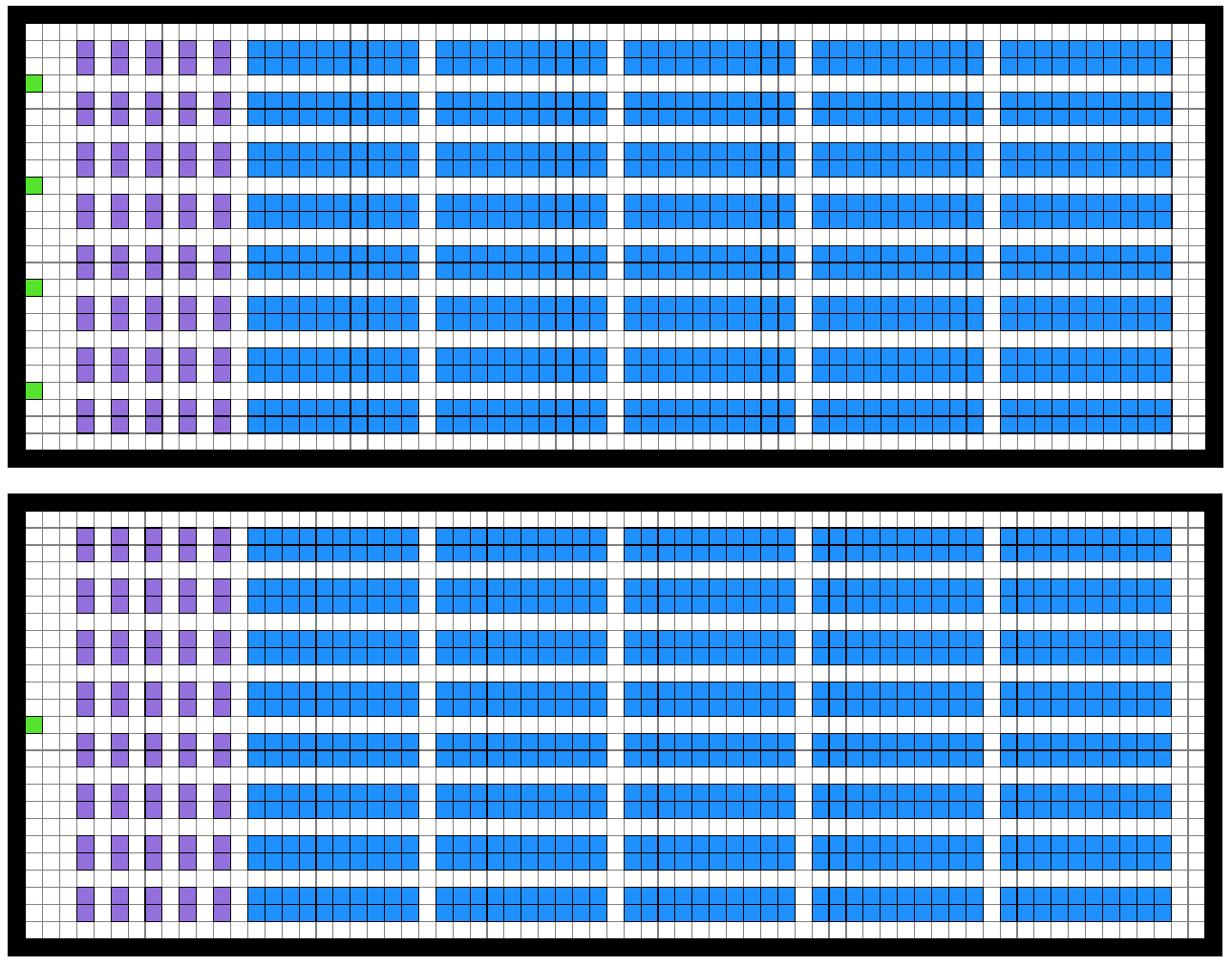} % Reduce the figure size so that it is slightly narrower than the column.
\caption{Warehouse maps: (1) Blue grids represent shelves \(S\). (2) Purple grids represent caches \(C\). (3) Green grids represent unloading ports \(U\). The upper map has multiple ports, while a single port is in the bottom one. In the multi-port map, each unloading port is associated with its own independent cache area, task queue, and set of agents. The cache areas are located near the unloading ports, within a range of  \(\pm 2\) rows. In contrast, the single-port map allows the port to utilize all available agents and caches. Given that the number of cache grids can influence the cache hit rate, we tested different configurations by varying the number of cache grids from 80 to 16. This was achieved by progressively removing cache grids column by column from right to left.
}
\label{fig:cal_mapf}
\end{figure}

Besides general L-MAPF definition, in this paper there are some special constraints on the problem. We primarily focus on 2D warehouse layout maps.
As shown in ~\Cref{fig:cal_mapf}, we categorize all non-obstacle map grids into four types: blue grids \(B=\{b_i | b_i \in V\}\) for shelves and each shelf stores a unique type of item identified by a type number \(idx \in \mathbb{N}\), white grids \(W=\{w_i | w_i \in V\}\) for normal aisle grids, green grids \(U=\{u_i | u_i \in V\}\) for unloading ports where agents deliver items to complete a task, and purple grids \( C=\{c_i | c_i \in V\} \) for cache grids which will be explained later. Instead of a target queue \(T\), there is a task queue \(Q = [q_1, q_2, ...]\). For each task \(q_i = (idx, u_{i}) \in Q\), where \(idx\) represents the type number of the item to be delivered and \(u_i \in U\) indicates the unloading port where the item should be delivered. \ral{The maximum number of items that one agent can hold is \(P\). We believe this setting is feasible for real robots such as Megvii MegBot-E Robot or Geekplus RS Robot.}

There is a task assigner (TA), external to the MAPF algorithm, that determines the specific target locations for agents based on tasks from \( Q \). 
In this paper, the TA assigns new target locations to idle agents based on the most recent tasks as they arrive at \( Q \), handling them one by one.
Each agent \( i \in I \) starts from \( s_i \in V \), and its target location is determined by the TA. The actions available to each agent include waiting in place or moving to an adjacent vertex, with each action taking a unit of time. The path of an agent \(i\), denoted as \( p^i = [v_0^i, v_1^i, \ldots, v_{T_i}^i] \), records the sequence of vertices traversed by the agent from its start to target location. Once an agent completes its path to the assigned target, the TA immediately assigns it a new target location.

\section{Related Work}

\subsection{Multi-Agent Path Finding (MAPF)}

(One-Shot) MAPF, which has been proved an NP-hard problem with optimality~\cite{yu2013structure}, has a long history~\cite{silver2005cooperative}. This problem is finding collision-free paths for multiple agents while minimizing a given cost function. It has inspired a wide range of solutions for its related challenges. 
Decoupled strategies, as outlined in \cite{silver2005cooperative,wang2008fast,luna2011push}, approach the problem by independently planning paths for each agent before integrating these paths. 
In contrast, coupled approaches \cite{standley2010finding,standley2011complete} devise a unified plan for all agents simultaneously. There also exist dynamically coupled methods~\cite{sharon2015conflict,wagner2015subdimensional} that consider agents planned independently at first and then together only when needed in order to resolve agent-agent conflicts. 
Among these, Conflict-Based Search (CBS) algorithm \cite{sharon2015conflict} stands out as a centralized and optimal method for MAPF, with several bounded-suboptimal variants such as ECBS~\cite{barer2014suboptimal} and EECBS~\cite{li2021eecbs}. Some suboptimal MAPF algorithms, such as Prioritized Planning (PP)~\cite{erdmann1987multiple,silver2005cooperative}, PBS~\cite{ma2019searching}, LaCAM~\cite{okumura2023lacam} and their variant methods~\cite{chan2023greedy,li2022mapf,okumura2023lacam3} exhibit better scalability and efficiency.

\subsection{Lifelong MAPF}

Compared to the MAPF problem, Lifelong MAPF (L-MAPF) continuously assigns new target locations to agents once they have reached their current targets. Agents do not need to arrive at their targets simultaneously in L-MAPF. There are mainly three types of ways to solve L-MAPF: solving the problem as a whole~\cite{nguyen2019generalized}, using MAPF methods but replanning all paths at each specified timestep~\cite{li2021lifelong,okumura2022priority}, and replanning only when agents reach their current target locations and are assigned new targets~\cite{vcap2015complete,grenouilleau2019multi,okumura2023lacam,okumura2023lacam3}. There are also algorithms that consider the offline setting in the L-MAPF scenario, where all tasks are known in advance, such as CBSS~\cite{ren23cbss}, which uses Traveling Salesman Problem (TSP) methods to plan task orders, and \cite{brown2020optimal}, a four-level hierarchical planning algorithm with MILP and CBS. However, in this paper we focus on an online setting where incoming tasks are not known in advance.

% \subsection{Cache}

% The cache~\cite{burks1946preliminary} serves as a crucial component in computer science designed to temporarily store data, enhancing the efficiency of future request processing. Its main goal is to speed up access to frequently used data, thus minimizing dependence on slower storage disks. Popular caching policies include Least Recently Used (LRU)~\cite{mccabe1965serial} and First-In-First-Out (FIFO)~\cite{King71a}, with studies indicating LRU's superiority over FIFO~\cite{albers2002paging,dan1990approximate,chrobak1999lru}. The caching concept is widely applied in various fields, such as Databases~\cite{altinel2003cache} and CDN~\cite{harchol1999resource}.

\subsection{Warehouse Storage Strategy}

Automated Storage and Retrieval Systems (AS/RS) have gained attention for their potential to enhance warehouse efficiency and reduce operational costs~\cite{GuGM07, RoodbergenV09}. Various strategies for assigning items to storage locations have been widely adopted and evaluated~\cite{RoodbergenV09,AzadehKR19}. The random storage policy allocates each item type to a randomly chosen storage location, offering high space utilization~\cite{park2001optimal}. The closest storage policy places new items at the nearest available storage location to minimize immediate travel distance~\cite{gagliardi2012storage}. The turnover-based storage policy assigns items to storage locations based on their demand frequency~\cite{li2020storage}.

The concept of utilizing temporal storage areas (caches) has been explored in automated warehouse design. Several studies focus on the optimal positioning of these caches~\cite{YU2008377, POHL2009367}, while others propose AS/RS design methods that quantitatively consider operational constraints, including cache size~\cite{OzakiHOHRO16}. However, existing works primarily focus on the high-level design and evaluation of warehouse layouts without addressing the low-level path planning component. Our work specifically studies and evaluates the use of caches within the context of L-MAPF, incorporating more detailed considerations of low-level path planning.

\section{Method}

In this section, we introduce our Lifelong MAPF with Cache Mechanism (L-MAPF-CM) framework, which includes a new type of map grid and a task assigner (TA) featuring a cache lock mechanism.

\subsection{Cache Grids}

As shown in \Cref{fig:cal_mapf}, purple map grids \(c_i \in C\) are caches, and these additional vertices serve as interim storage areas to reduce the travel time of agents retrieving items. We assume each shelf \(b_i \in B\) stores an unlimited supply of a unique item with type number $idx$. There will be a total of \(M\) different types of items corresponding to the number of shelves. The L-MAPF-CM framework can have multiple unloading ports, and we refer to each unloading port and its surrounding cache as a group. Each group operates independently in terms of its task and agents. Agents within a group can only accept tasks from that group and use the cache grids belonging to that group. Item in each task \(q_i \in Q\) must be delivered to the corresponding unloading port. 

To the best of our knowledge, no previous MAPF/L-MAPF works have explored cache mechanism with MAPF/L-MAPF problem. We adopt the following assumptions to fit the warehouse senerio:
\ral{(1) Agents have limited capacity to carry items and they are also limited to transporting one type of items at a time.}
\ral{(2) Each cache grid has the same capacity of agents and also are limited to store one type of items.}
\ral{(3) Items evicted from cache should be sent back to shelves.}

\subsection{Task Assigner (TA)}

The TA operates externally to the L-MAPF algorithm and has the ability to define all agents' target locations at any given timestep.
When the TA encounters a new task in $Q$ requiring completion, the first step typically determines where to get the item and which location (cache or shelf) should be allocated to an available agent. Since the design of a smart TA is not the focus of this paper, new tasks are assigned to available agents one by one. Generally, the TA first checks if the task item exists in cache grids. If it does, the TA assigns the cache location where the item is stored to an agent. If not, the TA assigns the item's shelf location to the agent, followed by a location in a cache grid where the items can be stored. Once the items are stored in the cache, the agent can transport one item to the unloading port.

\ral{However, similar to caching in computer architecture, there is a risk that other agents might replace the item in the cache with another item in multi-agent scenarios. For example, agent \(i\) wants to read items in cache \(c_k\), but items in \(c_k\) are replaced by agent \(j\) when \(i\) is going to \(c_k\). This situation could lead to L-MAPF algorithm, such as LaCAM, causing agents to move back and forth between cache and shelf grids. To mitigate this, a lock mechanism is employed, ensuring that agents can secure an item after confirming its availability in the cache grids.}

\subsection{Cache Lock Mechanism}

The cache lock mechanism is supported by a state machine to ensure efficient and synchronized access to cache locations. This approach prevents race conditions during cache interactions. Every cache grid will have its independent locks. Here we define reading and writing operations: reading refers to agents retrieving items from locations, while writing refers to agents placing items into locations. And readable refers to agents can retrieve items from locations, writable means agents can place items into locations. The cache lock mechanism ensures that no agent needs to return to the shelf when its target item is readable in the cache and that no two agents can access the same cache block concurrently for writing. This concurrency control is critical for preventing race conditions and maintaining cache consistency.

\subsubsection{Lock Types}
The cache lock mechanism is designed around three lock types, facilitating controlled access to the caches:
(1) \textbf{Read Lock (Shared Lock)}: This type of lock permits an agent to access the corresponding cache to retrieve items. The maximum number of lock shares is the number of items stored in the cache.
(2) \textbf{Write Lock (Exclusive Lock)}: This type of lock permits an agent to insert/remove items into/from the corresponding cache. Only one unique agent can hold a write lock of a cache.

\subsubsection{Lock Acquisition and Release}
The mechanism defines a protocol for lock acquisition and release:
(1) \textbf{Acquisition:} \textbf{Read Lock:} To obtain a read lock, an agent must verify that no write lock is currently held by any other agent on the cache. Multiple agents can simultaneously hold the read lock on a single cache. This arrangement ensures that when agents need to retrieve items from the cache, the item remains unchanged until all agents have successfully retrieved it when multiple identical tasks are assigned to different agents. \textbf{Write Lock:} To secure a write lock, an agent must confirm that the target cache location is free from any read or write locks. A write lock on a cache will be acquired in two situations: (a) when an agent needs to insert items into the cache, or (b) when an agent needs to remove items from the cache and return them to the shelf. Write locks are exclusive, implying that once an agent gets a write lock on a cache location, no other agent can acquire either a read lock or another write lock on that location. This ensures that an agent can insert/remove items without impacting other agents.
(2) \textbf{Release:} Agents will immediately release all locks they hold once they arrive at their target cache, as they have accomplished their intended tasks.

\begin{algorithm}[!t]
\small
\caption{L-MAPF-CM with One Group Overview}
\label{alg:cache_algo}
\textbf{Input}: Map \(G\), Initial Locations \(S\), Task Assigner TA, Agents \(agents\), Cache Grids \(caches\)
\begin{algorithmic}[1]
\State portLoc = getPortLoc($G$)
\For{agent in $agents$}
    \State agent.taskItem, agent.targetLoc = TA.getTask($G$)
    \State agent.curLoc = $s_i$
    \State agent.status = SF\_GET
\EndFor
\While{not TA.Q.empty()}
    \State plan = L-MAPF($G$, $agents$)
    \State excute(plan, $agents$)
    \For{agent in $agents$}
        \State TA.releaseLocks(agent, portLoc, $caches$)
    \EndFor
    \For{agent in $agents$}
        \State TA.updateStatusAndTargets(agent, portLoc, $caches$)
    \EndFor
\EndWhile
\end{algorithmic}
\end{algorithm}

\begin{figure}[t!]
\centering
\includegraphics[width=0.4\textwidth]{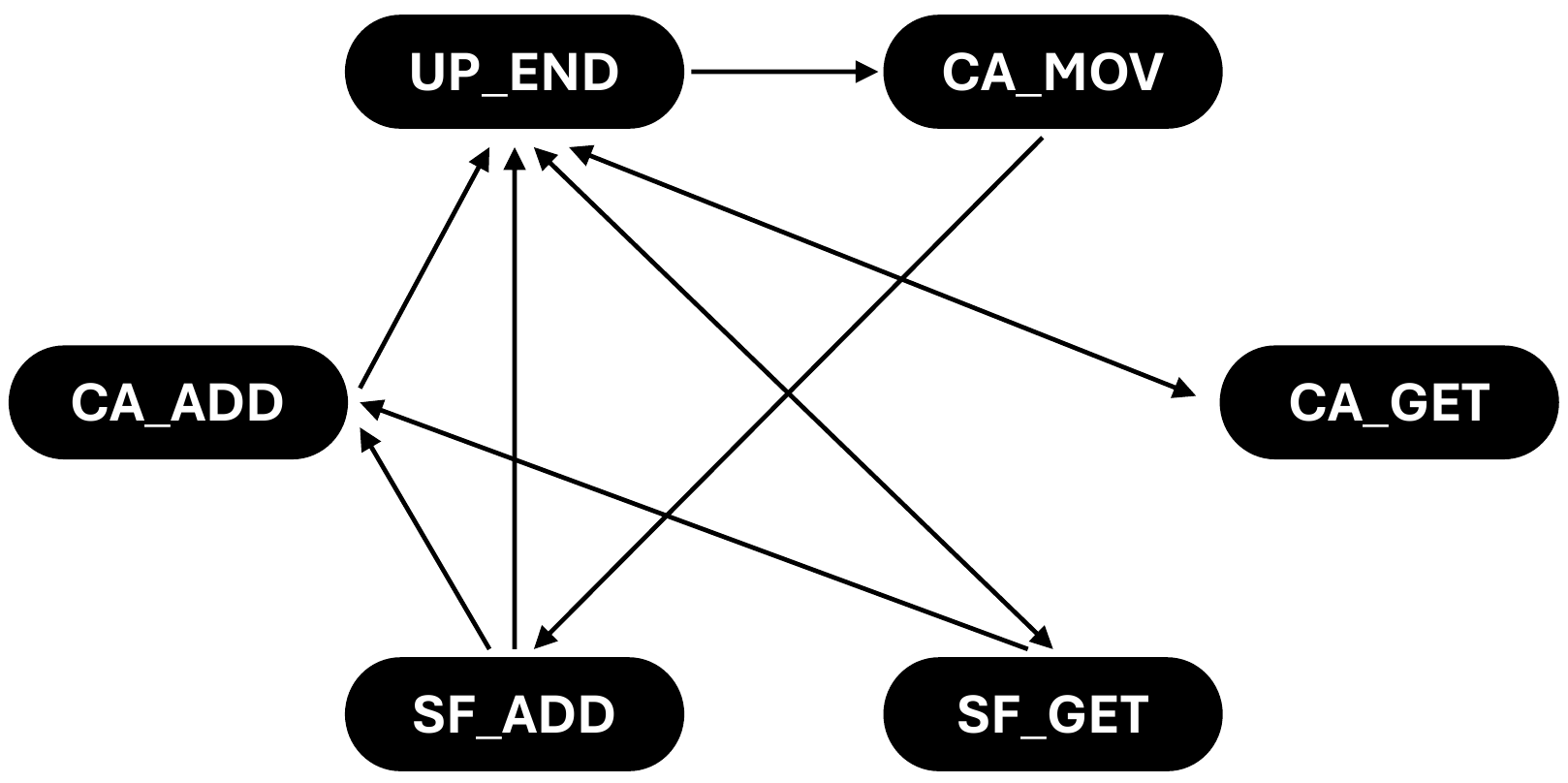} % Reduce the figure size so that it is slightly narrower than the column.
\caption{State Machine: (a) \textbf{SF\_GET}: The agent is moving to a shelf to retrieve its task item. (b) \textbf{CA\_MOV}: The agent is moving to a cache to remove all items. (c) \textbf{CA\_GET}: The agent is moving to a cache to retrieve a task item. (d) \textbf{CA\_ADD}: The agent is moving to a cache to store task items. (e) \textbf{SF\_ADD}: The agent is moving to a shelf to return items from a cache. (f) \textbf{UP\_END}: The agent is moving to its unloading port with one task item.}
\label{fig:state_machine}
\end{figure}

\begin{figure}[t!]
\begin{subfigure}[b]{0.15\textwidth}
\centering
\includegraphics[width=0.8\textwidth]{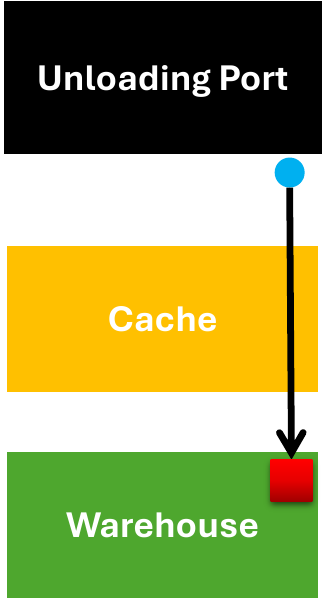} % Reduce the figure size so that it is slightly narrower than the column.
\caption{SF\_GET
}
\label{fig:status1}
\end{subfigure}
\begin{subfigure}[b]{0.15\textwidth}
\centering
\includegraphics[width=0.795\textwidth]{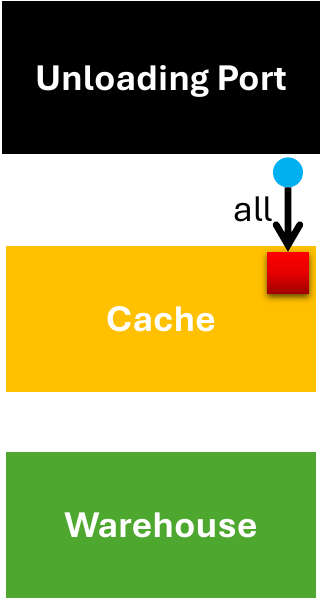} % Reduce the figure size so that it is slightly narrower than the column.
\caption{CA\_MOV
}
\label{fig:status2}
\end{subfigure}
\begin{subfigure}[b]{0.15\textwidth}
\centering
\includegraphics[width=0.8\textwidth]{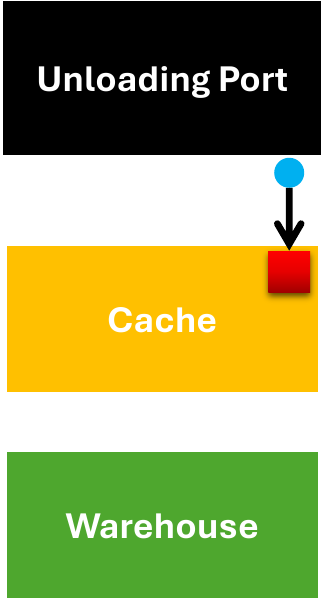} % Reduce the figure size so that it is slightly narrower than the column.
\caption{CA\_GET
}
\label{fig:status3}
\end{subfigure}

\begin{subfigure}[b]{0.15\textwidth}
\centering
\includegraphics[width=0.795\textwidth]{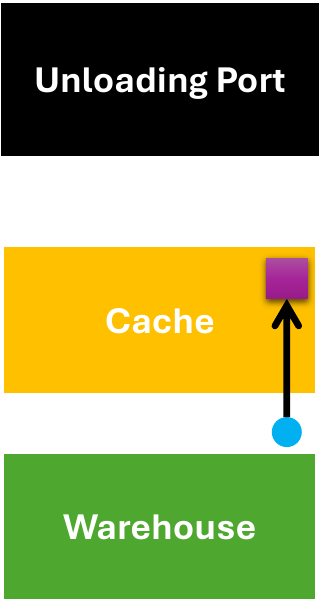} % Reduce the figure size so that it is slightly narrower than the column.
\caption{CA\_ADD
}
\label{fig:status4}
\end{subfigure}
\begin{subfigure}[b]{0.15\textwidth}
\centering
\includegraphics[width=0.8\textwidth]{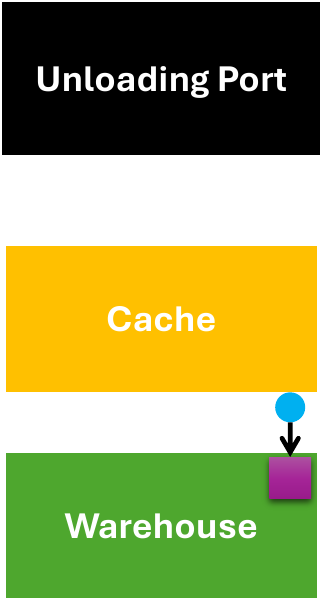} % Reduce the figure size so that it is slightly narrower than the column.
\caption{SF\_ADD
}
\label{fig:status5}
\end{subfigure}
\begin{subfigure}[b]{0.15\textwidth}
\centering
\includegraphics[width=0.8\textwidth]{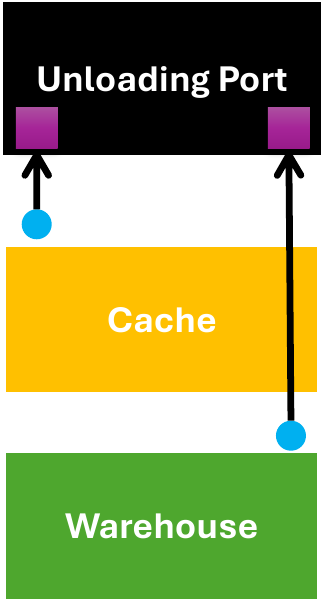} % Reduce the figure size so that it is slightly narrower than the column.
\caption{UP\_END
}
\label{fig:status6}
\end{subfigure}
\caption{Status examples: The blue circle represents agents, the purple square represents items to be placed, and the red square represents items to be taken out.}
\label{fig:status}
\end{figure}

\subsection{Algorithm}

\ral{\subsubsection{Overview}
As shown in \Cref{alg:cache_algo}, we introduce the entire procedure for a single group in L-MAPF-CM. L-MAPF-CM consists of three parts: initialization, path planning and execution, and status updates. }

\ral{\textbf{Initialization} (Lines 1-5): In this phase, all agents are assigned their task items, and their target locations are determined. Since the caches are empty at the beginning, agents can only retrieve items from the shelves. Therefore, the agents' initial target locations are the shelves, and their status is set to SF\_GET.}

\ral{\textbf{Path Planning and Execution} (Lines 7-8): Once all agents' target locations have been determined, we use any L-MAPF algorithm to generate a collision-free plan until at least one agent reaches its target. All agents then move according to the generated plan. After this step, at least one agent will have arrived at its target location. }

\ral{\textbf{Status Update} (Lines 9-12): In this phase, TA first releases all locks held by agents that have arrived at the caches, allowing other agents more flexibility in utilizing the caches. After releasing the locks, the TA determines the next target locations for all agents and updates their statuses accordingly. Note that each agent may receive a new target depending on its current status. With the new targets assigned, the L-MAPF algorithm is used again to generate a new collision-free plan. This process repeats until all tasks are completed.}

\ral{\subsubsection{State Machine}}

\ral{As shown in \Cref{fig:state_machine,fig:status}, we now discuss how TA maintains locks and agent status. We begin with the simplest status, \textbf{UP\_END}, where each agent carries one task item to the unloading port. Once the agent arrives at the unloading port, the TA assigns a new task item and decides the agent’s next target location. TA will first check if there is any readable cache. (a) One readable cache: the agent acquires a read lock on the cache and is set to CA\_GET. Next target is the cache. (b) No readable cache, but an empty cache exists: the agent acquires a write lock on the cache and is set to SF\_GET. Next target is the shelf. (c) No readable or empty cache, but a writable cache exists: the agent acquires the write lock, sets the target as the cache, and the status is updated to CA\_MOV. (d) No readable, writable or empty cache: the agent is set to SF\_GET. Next target is the shelf.}

\ral{When \textbf{CA\_GET} agents arrive at the cache, they release their read locks and retrieve one task item from the cache. Their target are then updated to the unloading port, and their status change to UP\_END. For each \textbf{SF\_GET} agent, the TA checks its status at every update. If a readable cache becomes available, the agent's target is updated to the cache, and its status is set to CA\_GET. When an SF\_GET agent arrives at its target location, the TA checks if the agent holds a write lock:
(a) Without a write lock: The agent takes one task item to the unloading port, and its status is updated to UP\_END. Next target is the unloading port.
(b) With a write lock: The agent’s status is updated to CA\_ADD, and it takes as many items as it can carry. Next target is the cache.}

\begin{figure*}[htbp]
\centering
\includegraphics[width=\textwidth]{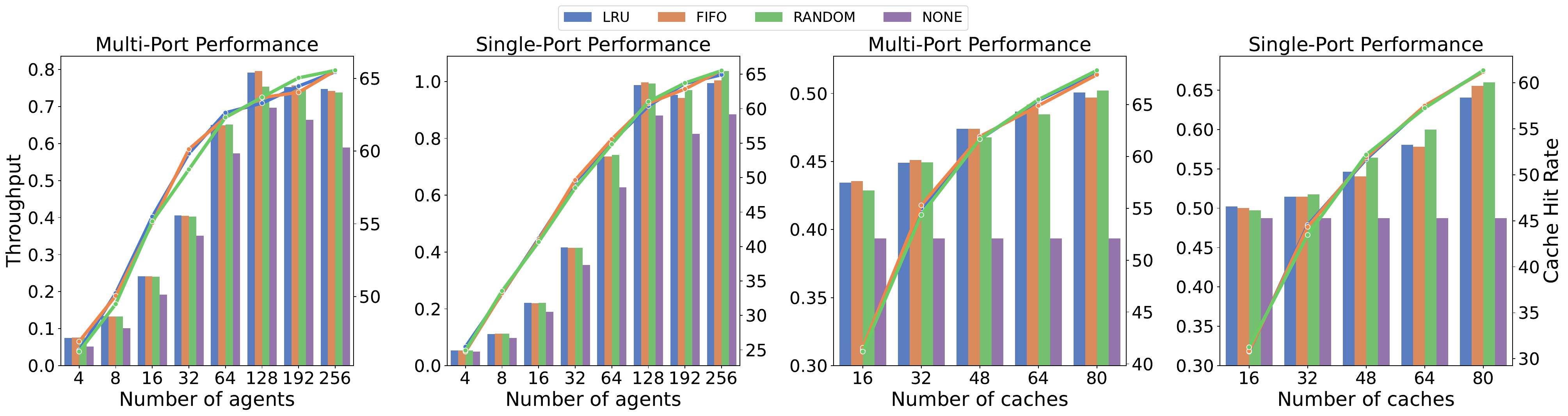} % Reduce the figure size so that it is slightly narrower than the column.
\caption{Throughput (Bar chart, higher is better) and Cache Hit Rate (Line chart, higher is better). LRU, FIFO, and RANDOM represent L-MAPF-CM with different cache replacement policies. NONE represents L-MAPF algorithm without cache. 
}
\label{fig:nagent_ncache_makespan_all}
\end{figure*}

\begin{figure*}[htbp]
\centering
\includegraphics[width=\textwidth]{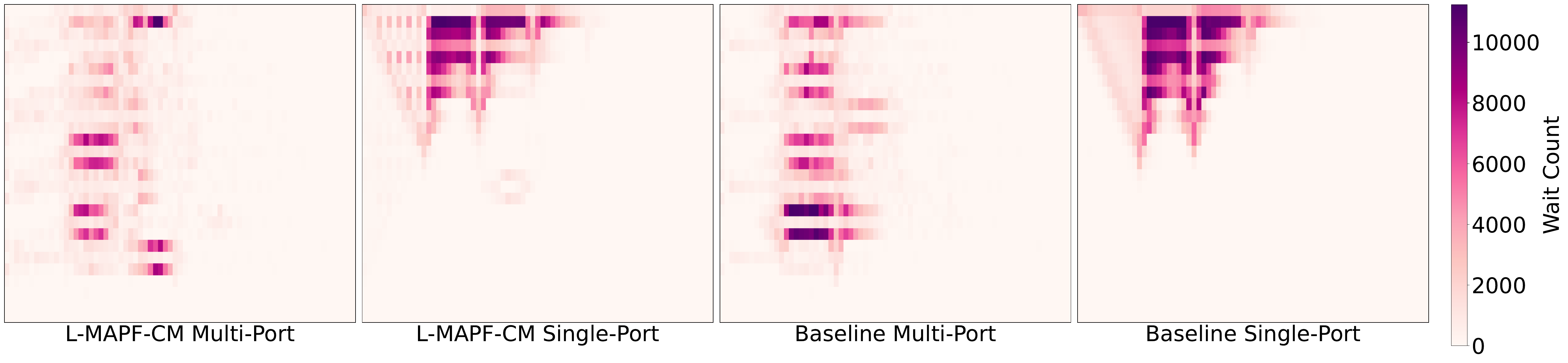} % Reduce the figure size so that it is slightly narrower than the column.
\caption{The frequency of agent wait actions on each map grid with 256 agents and 80 caches under Zhang distribution: Given the large number of agents, both L-MAPF-CM and the baseline experience significant traffic congestion (dark color area). The bias in congestion positions across the map is attributed to item indices and tasks in $Q$.}
\label{fig:WaitMaps}
\end{figure*}

\ral{When \textbf{CA\_MOV} agents arrive caches, they get all items from the cache and head to the shelf corresponding to those items. They release their write locks, and their status changes to SF\_ADD. When \textbf{CA\_ADD} agents arrive at the cache, they store the items except for one. Their target then becomes the unloading port, and their status is updated to  UP\_END. When \textbf{SF\_ADD} agents arrive at the shelf, they return all items to the shelf. Their status is then updated to SF\_GET, with their new target being the shelf containing their task items.}

\ral{We give out two examples. \textbf{(a) the target item is already in the cache}: 1) the TA assigns the agent the goal to the cache grid that holds the target item—the state status transfers from UP\_END to CA\_GET. 2) Upon the agent arriving at the cache grid, the target item is returned to the port, and the state status transfers from CA\_GET to UP\_END.
\textbf{(b) the target item is not in the cache}: 1) the cache is not full, we do not need to perform garbage collection. The TA assigns the agent the goal to the shelf grid that holds the target item—the state status transfer from UP\_END to SF\_GET. 2) Upon the agent arriving at the cache grid, we find an empty available cache grid, so the TA assigns the agent the goal to that cache grid, and the state status transfers from SF\_GET to CA\_ADD. 3) Then, the agent will insert the item into the cache grid and continue to bring the item to the port to finish the task; the state status transfers from CA\_ADD to UP\_END.  }

\ral{\subsubsection{Deadlock/Starvation Free and Correctness of State Machine}}

\ral{First, we show the lock algorithm is free from deadlock and starvation: no agents are always stuck in a circle waiting with other agents or waiting forever while accessing the cache. When an agent wants to read (shared lock) or write (exclusive lock) a lock of a cache grid, it is either successful or failed. If read fails, it will not hold the read lock or wait for a cache hit, but will go directly to the shelf to get that item. If write fails, it will not hold the write lock or wait for an empty cache grid, but directly deliver the item to the port. Thus, one of the conditions necessary to form a deadlock, Hold and Wait~\cite{coffman1971system}, is broken, so the deadlock can never occur. Additionally, each agent completes its operation in a bounded number of steps without infinite waiting for the resources, so we can conclude that this lock algorithm is free from starvation.}

\ral{Second, we show that state machine itself never gets stuck. This property equals the statement: there is no cycle in state machine without containing the status UP\_END: starting from status UP\_END, we can finally return to status UP\_END after finite steps of transformation. As shown in \Cref{fig:state_machine}, if we remove the status UP\_END and all corresponding directed edges, there is no cycle remaining in the graph. }

\section{Experimental Results}

We use L-MAPF without cache as a baseline, which aligns with our problem definition. To evaluate performance, we compare L-MAPF-CM with L-MAPF, both using LaCAM to generate collision-free paths. L-MAPF-CM and L-MAPF were implemented in C++, building on parts of the existing LaCAM codebase~\cite{okumura2023lacam}\footnote{Our code and video samples are available at supplementary material.}
All experiments were conducted on a system running Ubuntu 20.04.1, equipped with an AMD Ryzen 3990X 64-core CPU and 64GB RAM at 2133 MHz.

\subsection{Test Settings} 

\begin{figure*}[htb]
\centering
\includegraphics[width=\textwidth]{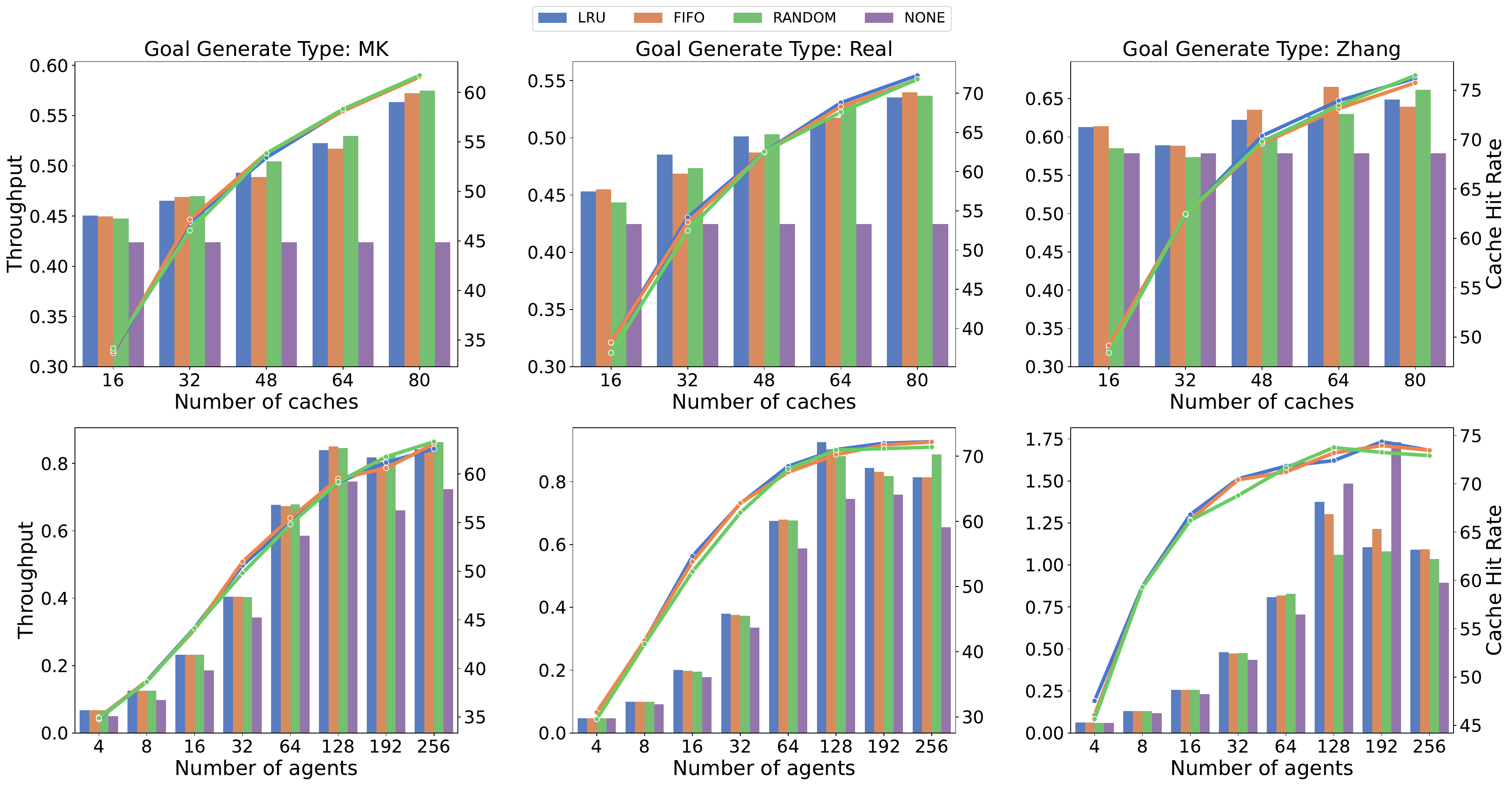} % Reduce the figure size so that it is slightly narrower than the column.
\caption{Throughput (Bar chart, higher is better) and Cache Hit Rate (Line chart, higher is better). As the number of caches increases, L-MAPF-CM's cache hit rate and throughput performance improve. In the MK, Real, and Zhang distributions, L-MAPF-CM surpasses the baseline in most test settings. However, it is observable that as the number of agents increases, the improvement offered by L-MAPF-CM diminishes. Additionally, an abnormal decrease in throughput is observed at baseline under the Zhang distribution with 256 agents.
}
\label{fig:nagent_makespan_all_goal}
\end{figure*}

\begin{figure}[t!]
% \begin{subfigure}[b]{0.48\textwidth}
\centering
\includegraphics[width=0.4\textwidth]{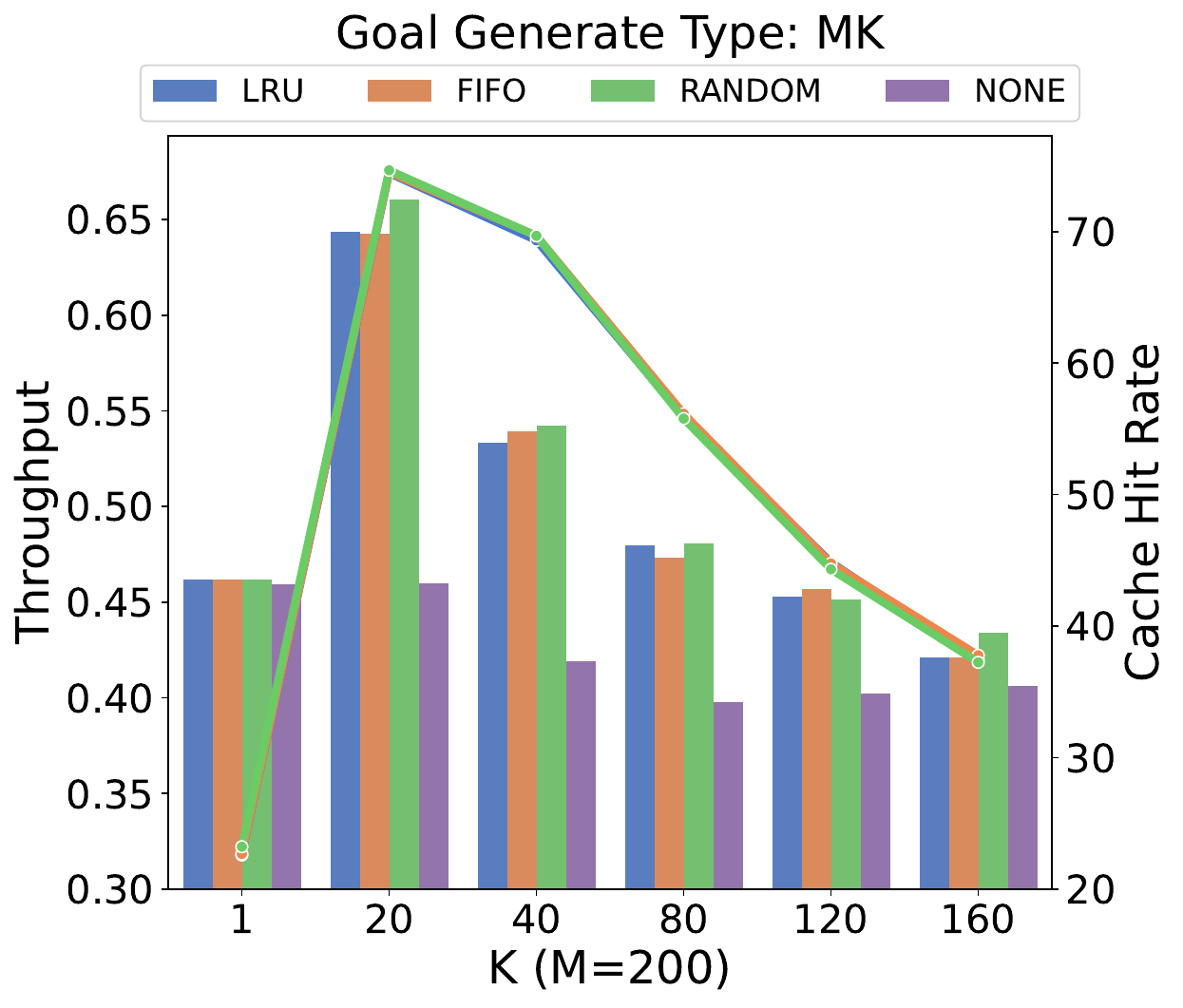} % Reduce the figure size so that it is slightly narrower than the column.
\caption{Throughput (Bar chart, higher is better) and Cache Hit Rate (Line chart, higher is better). The hit rate significantly depends on the input task distribution, and the hit rate also greatly affects the final throughput performance.
}
\label{fig:nk_makespan_MK}
% \end{subfigure}
\end{figure}

As shown in \Cref{fig:cal_mapf}, we demonstrated a warehouse map (27x71) with caches based on our problem definition. The map is adapted from warehouse map of MAPF benchmark~\cite{stern2019multi}. It includes 1600 shelf grids, a maximum of 80 cache grids and 4 unloading ports. The maximum cache-to-shelf ratio is 5\%. We tested L-MAPF-CM and baseline in both multi-port and single-port scenarios, as depicted in \Cref{fig:cal_mapf}. Since the design of a smart TA is not the focus of this paper, and a naive TA cannot handle multi-port scenarios effectively, we will use multiple groups to test scenarios with multiple unloading ports. The multi-port scenario has 4 working groups of unloading ports, cache grids and agents, each with a maximum of 20 cache grids. The single-port scenario maintains the same number of total cache grids and agents as the multi-port but only has one unloading port. Because each shelf grid represents a unique kind of item, we randomly assign an index to each shelf grid. We test all scenarios with different cache numbers \{16, 32, 48, 64, 80\} by deleting some cache grids (refer to \Cref{fig:cal_mapf}). We have also chosen various total numbers of agents, \(\{4, 8, 16, 32, 64, 128, 192, 256\}\). In the multi-port scenario, each group has an equal number of agents \(\{1, 2, 4, 8, 16, 32, 48, 64\}\). Each agent carrying capacity \(P\) is 100.  All data shown in \cref{fig:nagent_makespan_all_goal,fig:nagent_ncache_makespan_all,fig:nk_makespan_MK} represent average values across all variables not displayed on the figures.

Since we can expect the distribution of the task queue could significantly affect the performance of cache design, we designed three input task distributions to test L-MAPF-CM, including: 
(1) $M$-$K$ distribution (MK): For any consecutive subarray of length $M$ in the task queue $Q$, there are at most $K$ different kinds of items. This distribution is inspired by~\cite{albers2002paging}, where LRU has been proven to have an upper bound on the cache miss rate and to be better than FIFO. This distribution can also represent several items people purchase daily, and some items may become very popular at one time, replacing previously popular items. 
(2) 7:2:1 distribution (Zhang): There are 70\% kinds of items with only a 10\% appearance probability in the task queue, 20\% kinds of items with a 20\% probability, and 10\% with a 70\% appearance probability~\cite{zhang2016correlated}. 
(3) Real Data Distribution (RDD): We obtain data from Kaggle's public warehouse data\footnote{kaggle.com/datasets/felixzhao/productdemandforecasting}, build a probability distribution based on the frequency of data and generate tasks from this probability distribution. 

We randomly generate agent start locations and \(Q\) for all input task distributions. We will continue generating tasks until 1,000 tasks have been completed. For the MK distribution, we select \(M=200\) and \(K\) values of \(\{1,20,40,80,120,160\}\). We allocate a total of 10 seconds for the L-MAPF solver to find a collision-free solution for all 1,000 tasks. Based on our algorithm, the runtime overhead of L-MAPF-CM is negligible, with the majority of the time being spent on the L-MAPF solver. We use throughput to show performance. For L-MAPF-CM, we have tested three different cache replacement policies. Least Recently Used (LRU), First-In-First-Out (FIFO), and RANDOM. We use NONE to represent the baseline method.

\subsection{Performance} 

Many variables can influence the performance of L-MAPF-CM, including the number of agents, the number of caches, and the distributions of input tasks. As illustrated in \Cref{fig:nagent_ncache_makespan_all}, increasing the number of caches improves cache hit rate and throughput performance for both single- and multi-port scenarios. It is also observed that as the number of agents continues to increase (from 64 to 256), there is not much difference in the throughput for both L-MAPF-CM and the baseline. As shown in \Cref{fig:nagent_makespan_all_goal,fig:nagent_ncache_makespan_all,fig:nk_makespan_MK}, the cache replacement policy won't affect cache hit rate and final throughput performance too much. 

In \Cref{fig:nagent_makespan_all_goal} and \Cref{fig:nk_makespan_MK}, we present the performance of L-MAPF-CM and the baseline for various task distributions. \Cref{fig:nagent_makespan_all_goal} demonstrates that as the number of caches increases, L-MAPF-CM's performance improves. 
In the MK, Real, and Zhang distributions, L-MAPF-CM surpasses the baseline in most test settings. However, it is observable that as the number of agents increases, the improvement offered by L-MAPF-CM diminishes.
As depicted in \Cref{fig:nk_makespan_MK}, the hit rate significantly depends on the input task distribution, and the hit rate also significantly affects the final throughput performance. When hit rates decrease, throughput also decreases.

We can also notice there are some sudden decreases of throughput in subfigures of \Cref{fig:nagent_ncache_makespan_all} and \Cref{fig:nagent_makespan_all_goal}. For example, an abnormal decrease in throughput is observed in the baseline under the Zhang distribution with 256 agents. One possible reason for this scenario is traffic congestion occurring on the map as the number of agents increases. Upon reviewing the test results, we found that this issue is caused by LaCAM. Since there are too many agents in the same small area of the map, LaCAM has a very low chance of producing a good solution unless all agents simply wait. Traffic congestion largely impacts low-level path planning part. \Cref{fig:WaitMaps} illustrates the frequency of agent wait actions on each grid when the number of agents reaches 256 with the Zhang distribution. Both L-MAPF-CM and the baseline experience severe traffic congestion in this scenario.

\ral{High cache hit rates and smooth traffic are crucial for the performance of L-MAPF-CM. As shown in \Cref{fig:nagent_ncache_makespan_all}, we can increase the number of caches to enhance the cache hit rate, and we can also directly add more agents to improve the throughput performance. However, both methods come with their drawbacks. The number of caches is constrained by the space available close to the unloading port. Introducing more agents can lead to severe traffic congestion, ultimately causing the final throughput to drop. Nevertheless, there are potential solutions to mitigate these adverse effects, such as implementing more efficient map designs~\cite{ijcai2023p611} and using one-way systems near the cache and unloading ports~\cite{zhang2024guidance}. Additionally, as we currently employ a simple task assigner, we could implement a more advanced policy with predictive capabilities, leveraging real warehouse task data to enhance the cache hit rate. The cache replacement policies, such as LRU and FIFO, may be also overly simplistic for L-MAPF-CM.Performance could also be improved with a more fine-grained cache locking algorithm. Incorporating more complex policies, including learning-based approaches, could further improve the cache hit rate, especially since L-MAPF-CM allows more planning time than traditional caches in computer architecture.}

\section{Conclusion}

This work presents L-MAPF-CM designed to improve the performance of L-MAPF. We have introduced a new map grid type called cache for temporary item storage and replacement. Additionally, we devised a locking mechanism for caches to enhance the stability of the planning solution. This cache mechanism was evaluated using various cache replacement policies and a range of input task distributions. L-MAPF-CM demonstrated performance improvements in most of the test settings. We also identified that high cache hit rates and smooth traffic are crucial for the performance of L-MAPF-CM. \ral{Therefore, there are many interesting avenues for future work, such as developing a smart TA to manage task order, creating a data-driven cache replacement policy to improve cache hit rate, implementing a traffic congestion avoidance method and designing a hierarchical cache system.}

% \section*{Acknowledgement}

\bibliographystyle{IEEEtran} % use IEEEtran.bst style
\bibliography{strings,myref}

\end{document}